# Spotting Separator Points at Line Terminals in Compressed Document Images for Text-line Segmentation

Amarnath R.
Department of Studies in Computer Science,
University of Mysore, India.

P. Nagabhushan
Department of Studies in Computer Science,
University of Mysore, India.

## ABSTRACT
Line separators are used to segregate text-lines from one another in document image analysis. Finding the separator points at every line terminal in a document image would enable text-line segmentation. In particular, identifying the separators in handwritten text could be a thrilling exercise. Obviously it would be challenging to perform this in the compressed version of a document image and that is the proposed objective in this research. Such an effort would prevent the computational burden of decompressing a document for text-line segmentation. Since document images are generally compressed using run length encoding (RLE) technique as per the CCITT standards, the first column in the RLE will be a white column. The value (depth) in the white column is very low when a particular line is a text line and the depth could be larger at the point of text line separation. A longer consecutive sequence of such larger depth should indicate the gap between the text lines, which provides the separator region. In case of over separation and under separation issues, corrective actions such as deletion and insertion are suggested respectively. An extensive experimentation is conducted on the compressed images of the benchmark datasets of ICDAR13 and Alireza et al [17] to demonstrate the efficacy.

## General Terms
Line separator points at every line terminal in a compressed handwritten document images enabling text line segmentation.

## 1. INTRODUCTION
Generally, a document image is represented in a compressed format. The format is developed based on the guidelines of CCITT Group 4 standards, which is a part of ITU (International Telegraph Union) [1]. This compression standard facilitates both efficient storage and transmission [12] and therefore it is utilized in real-time applications including fax machines, photocopy machines, digital libraries and communication networks.

The compressed representation of a document could imply a solution to the big data problems arising from the document images, particularly with regard to storage and transmission. However to perform digital document analysis (DDA) [20], the image in the compressed format has to undergo the decompression stage [13, 14. 15]. This pre-requisite warrants additional buffer space and also extra time. If DDA could be carried out directly in the compressed version, then the document image compression could be viewed as an effective solution to the big data problem arising from the document images.

Few literature reported working directly on the compressed version of the printed text. But the challenging job is to perform DDA on the handwritten images because of oscillatory variations, inclined orientations and frequent touching of text lines while scribing the text lines. Therefore, performing the segmentation in an uncompressed handwritten text could be a difficult task. However, we foresee this possibility in view of the literature presented by Javed et al [13] which reports the research effort in the compressed printed document. In this research paper, the proposal is to spot the separator points at every line terminal in the compressed handwritten images enabling text line segmentation.

The CCITT Group-3 / Group-4 and JBIG protocols are developed based on the run-length encoding (RLE) [15], widely accepted for binary document compression. The RLE of the document image is represented in a matrix format. The first column of the matrix always starts with a white space. This represents the left margin. The depth (length) of the white space is larger at the separation points compared to the depth of the white space when it encounters the text lines. However, the last column of RLE does not infer the depth of right margin of the document. So we create a virtual column containing last non-zero value of every row of the RLE data. In summary, we propose to make use of a single column in-order to find the separator points at every line terminal.

If the depth of the white space in the first column is equal to that of the document width, then it certainly infers a text-line segmentation. But most of the time this situation may not occur in case of handwritten texts. This is because of oscillatory variations, inclined orientations and frequent touching of consecutive lines, particularly while writing on the white page (un-ruled paper). Further, every text-line in a handwritten document does not necessarily start or end with same left/right margin space. Such situation is more evident in the first and the last text-lines in every paragraph.

The depth of the white space would be larger at the point of separation than the depth elsewhere, but this depth may not be pronounced even at the expected separator points when two text lines are touching at the beginning of the line itself and thus it causes under separation. So the correction is to insert a separator points. Similarly a deeper margin for consecutive text lines could also cause under separation, by showing the entire stretch as one separation. On the other hand, a larger concavity in the character, a higher indent space, and disjoint composition of a character may result in a perceivably high depth and hence a pseudo separation, causing over separation. Therefore, the correction requires the deletion of such over separation points.

The organisation of the paper is as follows – Section 2 contains related research work. Section 3 includes an understanding of the RLE structure. The algorithmic model of the proposed method is explained in Section 4. Experimental analysis conducted on benchmark datasets is presented in Section 5. Summary and future possibilities are presented in Section 6.

## 2. RELATED WORK

In spite of the extensive search, there is no contribution reported in the field of DDA directly operating we could identify some related works pertaining to the compressed image processing. Most of the contributions are in the field of skew detection / correction, document matching and archival. The overview of this literature is covered in Table 1. All the literature papers presented in this table refer to some DDA on the compressed printed document images.

In summary, the motivation is the absence of the work on the compressed version of handwritten document, and the hope that can be traced particularly because of [8,13,14].

## 3. COMPRESSED IMAGE REPRESENTATION AND TERMINOLOGIES

The CCITT Group 3 [2] or Modified Huffman (MH) [15] image format primarily uses line by line coding technique. Basically the MH uses RLE as its basis encoding function. RLE describes the length of the run that carries similar pixel value which is either 0 or 1. The pixel carrying value 1 (on) is interpreted as foreground whereas the pixel carrying the value 0 (off) is considered as background. An example of RLE format is represented in the table 2.

The RLE consists of alternate columns of number of runs of 0 and 1 acknowledged as odd columns (1, 3, 5,…) and even columns (2, 4, 6,…) respectively. The column always starts with white runs. In absentia of a white run at the starting point that is in the first column, it is essential to make an entry as 0 (note the line 7 and 8 in Table 2). Further this table shows how the RLE compression technique is involved in shrinking a binary image data of length say 14 bits to 5 columns. Each value in the RLE represents the magnitude or depth of the corresponding runs.

**Table 1. Related research work**

| Research Area | Authors | Contribution |
|---|---|---|
| Skew detection / correction in CCITT Group 4 | Shulan Deng et al [3] | Exploiting 2-dimensional correlation between scan lines by extracting connected component. Employed occurrence frequency of word objects |
| Skew detection on Run data | Y. Shima et al [5] | Coordinate transformation based on projection profile method. |
| Skew detection Directly on compressed CCITT Group 4 | A.L. Spitz [6] | Used position locations of black and white structures to determine skew angles. |
| Skew detection in JBIG | J. Kanai et al [7] | Used projection profile for predicting skews |
| Object Identification | C. Maa [4] | Attempted in identifying a bar code directly in compressed CCITT Group 4 images. A particular pattern from relative position of pixels between scan lines were used. |
| Layout Analysis | E. Regentova et al [9] | Used the connected-component-detection and labelling techniques on JBIG-encoded images for obtaining global layout |
| Document Retrieval | J. J. Hull [10, 11] | Used passcode of CCITT Group 4 as feature vectors. He used Hausdorff distance measure for document matching |
| Document Retrieval | Yue Lu et al [12] | Have worked on connected component techniques of CCITT Group 4 standard images. Word objects are bounded by extracting changing elements. These word objects are matched based on weighted Hausdorff distance |
| Segmentation | Mohammed Javed et al [8, 13, 14] | Have performed Line, Word, and Character Segments directly from run-length compressed data. They have used horizontal projection profiled and local minima points to estimate the text lines. |

**Table 2. Binary image data [13]**

| Line | Binary data | 1 | 2 | 3 | 4 | 5 |
|---|---|---|---|---|---|---|
| 1 | 00000000000000 | 14 | 0 | 0 | 0 | 0 |
| 2 | 00110000111110 | 2 | 2 | 4 | 5 | 1 |
| 3 | 01111000111110 | 1 | 4 | 3 | 5 | 1 |
| 4 | 01111000111110 | 1 | 4 | 3 | 5 | 1 |
| 5 | 01111000111110 | 1 | 4 | 3 | 5 | 1 |
| 6 | 00110000000000 | 2 | 2 | 10 | 0 | 0 |
| 7 | 10000000000000 | 0 | 1 | 13 | 0 | 0 |
| 8 | 10000000000000 | 0 | 1 | 13 | 0 | 0 |
| 9 | 00100001111100 | 2 | 1 | 4 | 5 | 2 |
| 10 | 01110001111100 | 1 | 3 | 3 | 5 | 2 |
| 11 | 01111001111100 | 1 | 4 | 2 | 5 | 2 |
| 12 | 01111100000000 | 1 | 5 | 8 | 0 | 0 |
| 13 | 00000000000000 | 14 | 0 | 0 | 0 | 0 |

For a better understanding, figure 1 (a) and (b) show a portion of the sample document image and its compressed version. Fig. 2 shows the RLE structure of this sample image.

**(a) Uncompressed document   (b) Compressed document**

**Fig. 1: Length pattern observed from a compressed text line**
**(Reference: A portion of ICDAR13 test image -214.tif)**

**Fig. 2: The RLE Structure**
**(Reference: A portion of ICDAR13 test image -214.tif)**

## 3.1 Depth of the White Space

The values in the first column of RLE represent the depth of the white space starting from the left border of the document page. Fig 3 shows the depth projection for a portion of the first column extracted from figure 2. It is observed that the entries in the first column are non-zero and this indicates a minimum white space as the left margin, even in the presence of the text-line.

**Fig. 3: The depth of the white space from left end of the document**
**(a) The first column of the RLE, (b) Projection of values**

The first column of the RLE implies the left margin of the document, whereas in case of the right margin the depth of the white space has to be traced in the RLE because it is not available as a column. Here, the last non-zero entry of every row of RLE is considered as the right margin of the document and hence a virtual column is built.

An illustration is provided in fig 4 where the last non-zero entries are taken from the odd column of every row. In some cases, the last non-zero entry appears in an even column and so a zero entry should be added for the virtual column of the corresponding row. A last non-zero entry in the even column indicates that the text-line touches the right border of the document.

**Fig. 4: The depth of the white space from right end**
**(a)   RLE format, (b) Virtual column**

## 3.2 Under Separation

One of the reasons for under separation is the touching or overlapping of two text-lines at the starting point itself. Here, the depth of the white-run is reasonably low. Fig 5 (a) shows an example where the text-lines 3 and 4 are touching each other at the beginning of the text line. The other reason is when a large margin space is indented at the beginning of the text line. This is shown in Fig 5 (b) where the second text-line has more left margin white space compared to other text-lines.

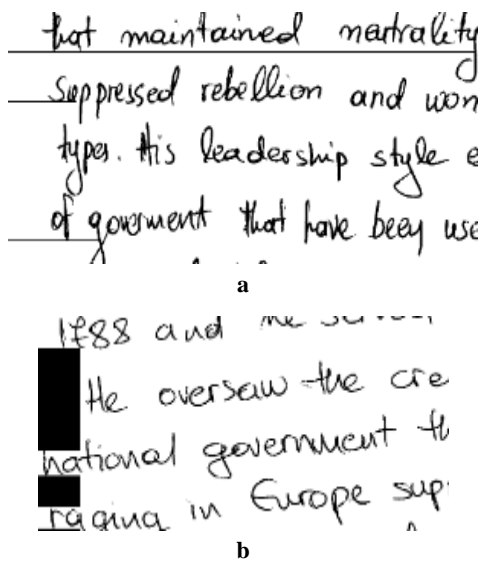

a

b

**Fig. 5: Under Separation**

**(Reference: A portion of ICDAR13 test images - 216.tif and 220.tif)**

**(a) Touching of lines at the starting point,
(b) Text line with more space for left margin**

## 3.3 Over Separation

The over separation occurs when a text line is identified as a non-text (white space) region. Fig 6 shows the character 'J' causing a pseudo separation point. The over separation is due to concavity of the character from the left end. The other affecting factor could be the multiple disjoint fractions or components which compose a character.

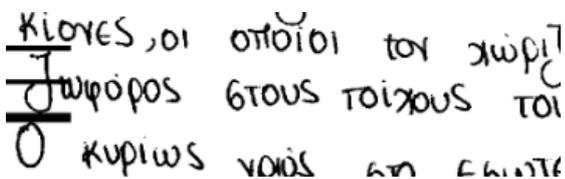

**Fig. 6: Over Separation
(Reference: A portion of ICDAR13 test image - 273.tif)**

## 4. IDENTIFICATION OF TEXT-LINE SEPARATORS IN COMPRESSED IMAGES

From the details presented in Section 3, there are three main stages – (a) Finding the bands of consecutive rows with larger white depths, (b) Finding the under separation (c) Finding the over separation. A detailed explanation is provided for each stage in the following sub sections.

## 4.1 Finding the bands of consecutive rows with larger white depths

The goal of this method is to identify the separator (non-text) and non-separator (text) regions. The first column of RLE and a threshold are the inputs. The threshold value (t) is heuristically chosen as 1/25 of the document width. This threshold is considered after analysing the other thresholds including 1/35 and 1/15 as well. Initially, we remove the margin space from the left border of the document by subtracting the values with a minimum value. After this elimination process, if the value is greater than the threshold, then the corresponding index position is labelled as separator point (say '1'), otherwise it is presumed as a text region (say '0'). Fig 7 (a) shows a sample image marked with separator bands (black patches) along the left border of the document. Fig 7 (b) shows the periodicity of the separator bands.

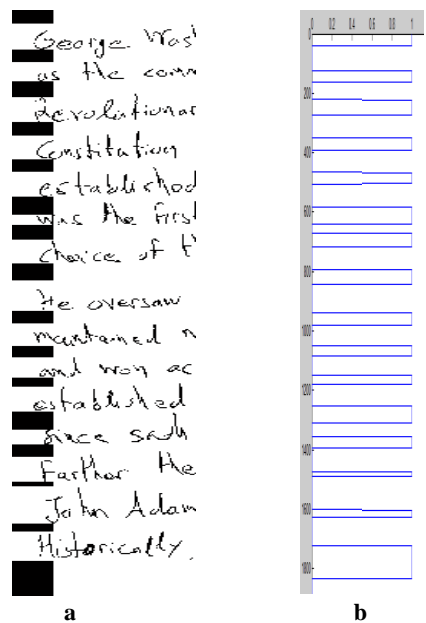

a b

**Fig. 7: Formation of Separator bands
(Reference: A ICDAR13 test image - 201.tif)**

**(a) Separator bands, (b) Periodicity of the separator band**

The first column of RLE and the threshold are represented as FC and t respectively in the algorithm. The final output, say Separator Band (SB), represents the region of text line separation.

**Algorithm** CreatingSeparatorBands

**Input:**  $FC - First\ Column\ of\ RLE\ data$
 $t - Threshold\ for\ the\ separator\ points$

**Output:**  $SB - Separator\ Bands$

**Step 1**: $Find\ minimum\ value\ from\ FC$

**Step 2**: $for\ i \leftarrow 1\ to\ size\ of\ FC$
  $SB(i) = FC(i) - minimum\ value$
  $SB(i) = \begin{cases} 1, & if\ SB(i) > t \\ 0, & otherwise \end{cases}$
  $else\ for$

**Step 3**: $Stop$

The time complexity of finding the minimum value is $O(m)$, where m = size of the first column. The algorithm scans the input array once again to find the separator points. Overall, the worst case of the algorithm is $O(2m)$.

## 4.2 Finding the under separation

The two factors causing under separation have been detailed in Section 3.2. In this section, we deal with the separator band width which is relatively large. The under separator region could be seen in Fig 8. Fig 5(b) in Section 3 shows the region of interest (ROI). When a separator band width is two times larger than that of an average band width, then it is presumed as under separator region or ROI. To resolve this, the ROI would be recursively iterated with the same algorithm described in the previous section. The recursion terminates when no ROI is detected.

On the other hand, the separator band width would be extremely large, sometimes it may cover more than 1/10 of the document height, which definitely affects the average separator band width. This scenario is shown in Fig 9. So we directly take this region as ROI and this would not be considered for calculating the average. The threshold 1/10 is chosen heuristically based on the average number of the text lines in the dataset.

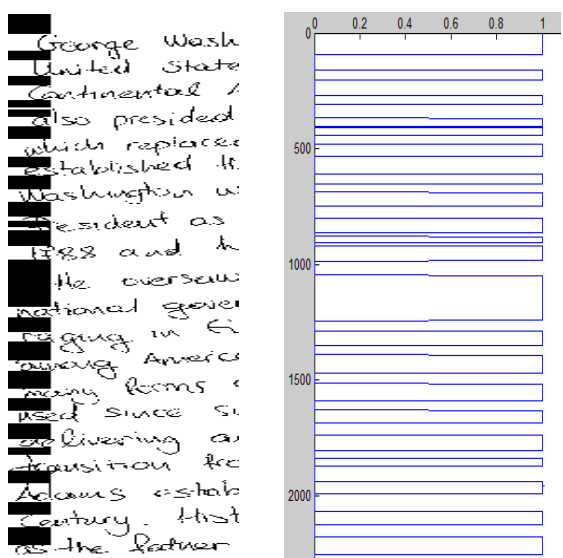

**Fig. 8: Separation band and frequency
(Reference: ICDAR13 test image - 220.tif)**

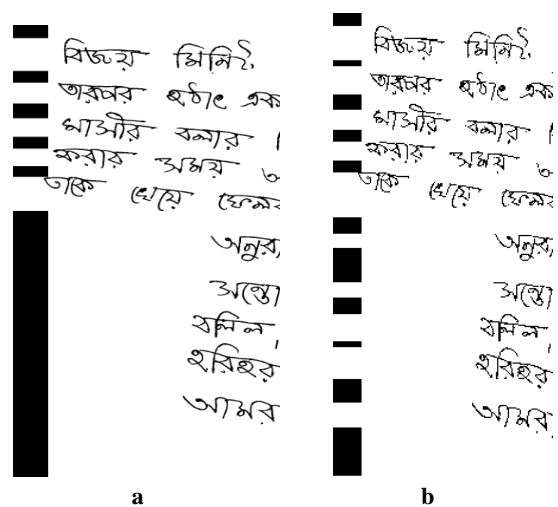

**a**    **b**

**Fig. 9: Separator bands
(Reference: A ICDAR13 test image - 201.tif)**

(a) A larger separation band,
(b) Separation bands after Iteration

**Algorithm** FindingUnderSeparation

**Input:** $SB$ – Separator Bands

**Output:** $ROI$ – Region of Interest –
Bands with two times larger
widths compared to the average

**Step 1**:

$for\ each\ separator\ band\ in\ SB$

$\quad if\ band\_width > \frac{1}{10}$

$ROI \leftarrow push(index\ position\ of\ the\ correspoding\ band)$

$\quad else$

$bandwidth\_sum \leftarrow bandwidth\_sum + band\_width$

$bandposition \leftarrow push(positions\ of\ the\ band\ \&\ width)$

$\quad end\ if$

**Step 2**:

$$averagebandwidth = \frac{bandwidth\_sum}{number\ of\ separator\ bands\ in\ bandposition}$$

**Step 3**:

$for\ each\ band\ in\ bandposition$

$\quad if\ band\_width > 2 \times averagebandwidth$

$ROI = push(index\ position\ of\ the\ correspoding\ band)$

$\quad end\ if$

**Step 4**: $Stop$

The time complexity for finding the average separator band width is $O(m), where\ m = size\ of\ the\ SB\ (input)$. The detection of the ROI is $O(m)$. The worst case scenario for this algorithm is $O(2m)$.

Next, the separator points are identified by taking the mid position of each band with respect to its position. Suppose the starting and the ending position of a separator band are $P_{start}\ and\ P_{end}$ respectively, then the mid-point is computed as $mid = \frac{P_{start}+P_{end}}{2}$. Fig. 10 shows the line separator example.

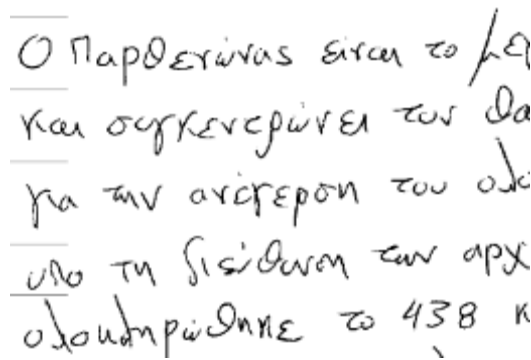

**Fig. 10: Line Separators
(Reference: A portion of ICDAR13 test image - 201.tif)**

The other under separation problem is illustrated in Section 3.2 (Fig 5(a)) when two adjacent text lines are touching at the beginning of the text-line. We analyze the frequency of the separator points. If the gap between the two adjacent separator points is more than twice its average gap, then it is considered as an under separation region. The under separation region could be seen in fig 11 with its corresponding separator frequency.

**Fig. 11: Under separation and frequency**
**(Reference: A portion of ICDAR13 test image - 214.tif)**

To resolve this, first we compute the average separation gap between the adjacent separators. Next, we re-compute the average separation gap by ignoring the touching separator points. This newly computed average is used in-order to insert the separator exactly in the midpoint of the two touching text-lines. The same algorithm to identify the under separation region is employed. Instead of computing the average separator band width, we take the average gap between the separator points. Therefore, the time complexity is $O(2m)$.

## 4.3 Finding the over separation
As described in Section 3.3, the reasons for over separation are disjoint character composition and perceivably higher concave character structure. The over separator points are detected based on the frequent appearance of the separator points than expected. This could be seen in Fig 11, where the separator line 6 is closely located to the separator 7.

**Fig. 12: Under separation and frequency**
**(Reference: A portion of ICDAR13 test image - 218.tif)**

The over separation points are detected when the gap between the adjacent is lesser than 1/3 of the average gap. In Fig 11, the gap between the separator points 6 and 7 is identified as over separation. In this scenario, the separator point 6 is to be removed because this separator point is comparatively closer to its adjacent point 5 than the gap between 7 and 8. The mathematical model is given below.

**Step 1**:

$$\text{AverageGap} = \frac{1}{\text{no of gaps}} \sum SB(\text{current}) - SB(\text{previous})$$

where current = current separator point

previous = previous separator point

**Step 2**:

For each separator gap

$$\text{if gap} < \frac{\text{AverageGap}}{3}$$

Remove the respective separator

The algorithm scans the separator points twice and so the overall time complexity is $O(2m)$.

## 4.4 Creation of a virtual column at the right end
To work on the right margin of the document image, we consider the last non-zero entry of every row of RLE data and we build a virtual column. This is explained clearly in Section 3.1. The algorithmic skeleton is provided here under.

**Algorithm** VirtualColumn

**Input:**   RLE

**Output:** VC – Virtual Column- consists of last non-zero value of every row in RLE data

**Step 1**: $for\ i \leftarrow 1\ to\ height\ of\ RLE$

    $for\ j \leftarrow width\ of\ RLE\ to\ 1$

    $if\ RLE(i,j) \neq 0$

        $VC(i) \leftarrow RLE(i,j)$

        $break$

    $endif$

    $end\ for$

    $end\ for$

**Step 2**: $Stop$

This algorithm takes $O(m \times n')$ where $m = row\ size\ of\ RLE\ and\ n' = column\ size\ of\ RLE$

Algorithms in 4.1, 4.2 and 4.3 can be applied on this virtual column to spot the separator points at right border of the document. A sample result of separator points at left and right border is shown in Fig 13.

45

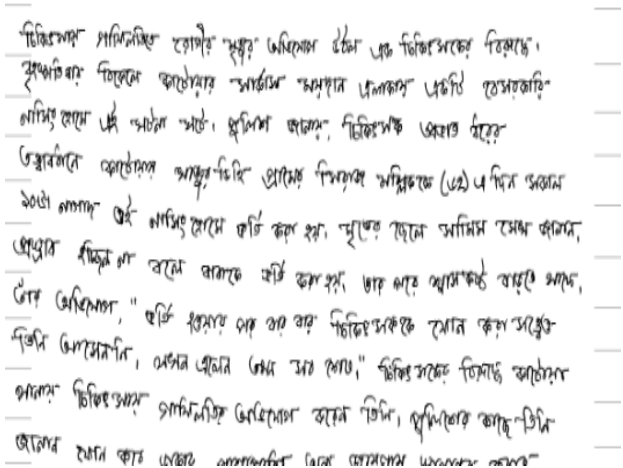

**Fig. 13: Separator points at left and right borders
(Reference: A portion of ICDAR13 test image - 305.tif)**

## 5. EXPERIMENTAL ANALYSIS

There is no standard compressed handwritten dataset available in the literature. However, the benchmark datasets such as ICDAR2013 [16], Alireza et al [17] of Kannada, Oriya, Persian and Bangla documents are compressed using the RLE technique. The compression standard is adopted as presented in [14]. The system is evaluated by counting the number of matches between the entities (separator points) detected by the algorithm and the entities present in the ground truth, proposed in the literature [16]. Let N be the count of ground-truth elements and the number of one-to-one matches be o2o, the detection rate (DR) is defined as follows:

$$DR = \frac{o2o}{N}$$

The machine learning statistics such as True Negative (TN) and False Positive (FP) in terms of under separation and over separation respectively is shown below.

$$TN = \frac{Under\ Separation\ Count}{Total\ Lines} \times 100$$

$$FP = \frac{Over\ Separation\ Count}{Total\ Lines} \times 100$$

The total separator points at left/right border of a document is the sum of the number of gaps between the text lines and the two margins (top and bottom) of the document page.

$$Total\ Separator\ points\ at\ \frac{left}{right}\ border\ of\ a\ document$$
$$= Total\ lines - 1 + 2 = Total\ Lines + 1$$

While experimenting, we ignore the separator point at the top margin of every document. The table 3 shows the DR on evaluating the algorithms on the handwritten datasets. The table shows one-to-one detection on both ends (left and right).

Different threshold values including 1/15 and 1/35 were experimented. However, the threshold value 1/25 would give relatively higher DR. In particular, the Persian handwritten dataset holds lesser DR. This is because the Persian characters or words are composed of disjoint components. For Persian texts the performance at the right end is better than left because it is written in left-to-right direction, causing a larger indent margin at left end when compared to its right.

**Table 3. Detection Rate tested with various compressed datasets**

| Datasets (Handwritten) | Total Lines (N) | Detected | | | | Undetected (%) | | | |
|---|---|---|---|---|---|---|---|---|---|
| | | o2o | | Rate (%) | | Left | | Right | |
| | | Left | Right | Left | Right | TN | FP | TN | FP |
| ICDAR13 [16] | 2649 | 2578 | 2502 | 97.31 | 94.45 | 2.69 | 2.78 | 5.55 | 6.44 |
| Kannada [17] | 4298 | 4173 | 4082 | 97.09 | 94.97 | 2.91 | 3.01 | 5.03 | 5.23 |
| Oriya [17] | 3108 | 3012 | 2911 | 96.91 | 93.66 | 3.09 | 4.10 | 6.34 | 7 |
| Bangla [17] | 4850 | 4650 | 4598 | 95.87 | 94.80 | 4.13 | 4.45 | 5.20 | 6.01 |
| Persia [17] | 1787 | 1690 | 1723 | 94.57 | 96.41 | 5.43 | 7.99 | 3.59 | 4.2 |

## 6. CONCLUSION AND FUTURE WORK

In this paper, a novel idea of working directly in the compressed representation of the document image is presented. We spotted the sequence of separator points at every line terminal in the RLE data. These separator points would enable the text line segmentation. Certainly, these points determine the text line segmentation in the printed compressed document. Though the entire RLE data is available, we used just the first column of the RLE to spot separator points on the left end of the document. In case of the right end, the last non-zero entry of every row in the RLE data is chosen to form a virtual column. The algorithm has some limitations in working with skews, large margins (indents), consecutive touching lines and disjoint characters. These limitations can be considered for the future work.